\pdfoutput=1

\documentclass[11pt]{article}

\usepackage[final]{acl}
\usepackage[table]{xcolor}
\usepackage{booktabs}
\usepackage{makecell}
\usepackage{arydshln}

\usepackage{times}
\usepackage{latexsym}

\usepackage[T1]{fontenc}

\usepackage[utf8]{inputenc}

\usepackage{microtype}

\usepackage{inconsolata}

\usepackage{graphicx}

\usepackage[most]{tcolorbox}

\usepackage{subcaption}
\usepackage{caption}
\usepackage{multirow}
\usepackage{xspace}

\title{The Cross-Lingual Cost: \\ Retrieval Biases in RAG over Arabic-English Corpora}

\author{\textbf{Chen Amiraz} \quad \textbf{Yaroslav Fyodorov} \quad \textbf{Elad Haramaty} \\
    \textbf{Zohar Karnin} \quad \textbf{Liane Lewin-Eytan} \\ \\
    Technology Innovation Institute \\
    \\
{\normalsize\texttt{\{chen.amiraz,yaroslav.fyodorov,elad.haramaty,zohar.karnin,liane.lewineytan\}@tii.ae}}}

\newcommand{\BGE}{{\em BGE-M3}\xspace}
\newcommand{\Efive}{{\em M-E5}\xspace}
\newcommand{\direct}{{\em direct}\xspace}
\newcommand{\oracle}{{\em language-oracle}\xspace}
\newcommand{\balanced}{{\em balanced}\xspace}
\newcommand{\translation}{{\em translation}\xspace}

\newtcolorbox[auto counter, number within=section]{promptbox}[2][]{%
  colback=gray!5,
  colframe=gray!80,
  fonttitle=\bfseries,
  title={Prompt~\thetcbcounter: #2}, 
  label={#1}, 
  boxrule=0.4mm,
  arc=2mm,
  outer arc=1mm,
  left=3mm,
  right=3mm,
  top=2mm,
  bottom=2mm,
  enhanced
}

\begin{document}
\maketitle
\begin{abstract}
Cross-lingual retrieval-augmented generation (RAG) is a critical capability for retrieving and generating answers across languages. Prior work in this context has mostly focused on generation and relied on benchmarks derived from open-domain sources, most notably Wikipedia. In such settings, retrieval challenges often remain hidden due to language imbalances, overlap with pretraining data, and memorized content. To address this gap, we study Arabic-English RAG in a domain-specific setting using benchmarks derived from real-world corporate datasets. 
Our benchmarks include all combinations of languages for the user query and the supporting document, drawn independently and uniformly at random. This enables a systematic study of multilingual retrieval behavior.

Our findings reveal that retrieval is a critical bottleneck in cross-lingual domain-specific scenarios, with substantial performance drops 
occurring when the user query and supporting document languages differ. A key insight is that these failures stem primarily from the retriever’s difficulty in ranking documents across languages. Finally, we propose two simple retrieval strategies that address this source of failure by enforcing equal retrieval from both languages or by translating the query, resulting in substantial improvements in cross-lingual and overall performance.  
These results highlight meaningful opportunities for improving multilingual retrieval, particularly in practical, real-world RAG applications. 

\end{abstract}

\section{Introduction} \label{sec:intro}

Retrieval-Augmented Generation (RAG) has emerged as the widely accepted approach for grounding large language models (LLMs) in external knowledge, with most research and development focused on high-resource languages, most notably English. However, many real-world applications, especially in corporate contexts, rely on multilingual corpora, where content spans both high- and low-resource languages. For example, internal knowledge management systems in governmental or legal domains often store content in both a high-resource language like English and the local language, while customer support systems may receive queries in the local language that require retrieving information from a corpus that mixes technical content in both languages. These scenarios introduce cross-lingual complexity, where users interact in a low-resource language while relevant information resides in a corpus containing documents in multiple languages. Prior work has shown that system performance in such cross-lingual settings tends to lag behind monolingual setups, due to challenges across both retrieval and generation \cite{wu2024not, sharma2025faux,park_investigating_2025}.  In this work, we focus on the bi-lingual English-Arabic setting -- a representative and important case of high- and low-resource language interaction.

Prior work has primarily focused on the generation component \cite{liu2025xrag, chirkova2024retrieval}, often using multilingual benchmarks derived from Wikipedia, the predominant open-domain source. However, evaluating retrieval in this context poses challenges due to several inherent characteristics: language imbalances, overlap with pretraining data, and the fact that much of Wikipedia’s knowledge is embedded in the model’s parametric memory. In contrast, our work focuses on the less explored retrieval component within a bilingual, domain-specific setting representative of real-world corporate applications. 
In this context, we study retrieval bias, namely the tendency of multilingual retrievers to favor one language over another, thereby overlooking relevant documents in the less-preferred language. In particular, we examine the cross-lingual setting, in which a query in one language may be answered by a document written in another. 

We construct benchmarks from UAE corporate datasets with parallel English-Arabic documents. Each benchmark includes a balanced set of English and Arabic queries, with answers grounded in a single language. The languages of the user query and supporting document are selected independently, enabling a systematic analysis of cross-lingual biases. Our analysis of these two benchmarks highlights that retrieval presents a significant bottleneck within the RAG pipeline. Moreover, the primary source of retrieval error arises in cross-lingual settings, namely when the query and the ground truth document are in different languages.

Finally, we propose two simple mitigation strategies tailored to the identified error source. The first strategy selects an equal number of passages from each language-specific subset, while the second searches the joint dataset twice, once with the original query and once with its translation. Both strategies result in substantial improvements in cross-lingual retrieval. The effectiveness of such basic interventions suggests that there remains considerable room for advancements in this area.

\section{Related Work} \label{sec:related}

Cross Lingual Information Retrieval (CLIR) is a critical capability for accessing knowledge across language boundaries, and has gained renewed attention with the rise of cross-lingual retrieval-augmented generation (RAG) systems. These systems typically operate in two phases, retrieval and answer generation. CLIR has historically been done via translation (see \citet{galuvsvcakova2021cross} and references within). With the rise of dense retrieval, most leading techniques avoid direct translation and instead embed queries and documents of different languages into the same space \cite{chen2024m3, louis2025colbert, wang2024multilingual, asai2021one}. The improved performance over retrieval tasks was also verified to occur in RAG for question answering by \citet{chirkova2024retrieval} that show an advantage to these direct methods over translation coupled with monolingual retrieval. The different retrieval techniques vary in their training method and data collections, yet all follow the same pattern of embedding the query and the document. They fall into the broader area of Cross Lingual Alignment, where the objective is to align representations of different languages \cite{hammerl2024understanding}. This broader area and the specifics of the different models are outside the scope of this paper.

For the answer generation phase, the challenge comes from the fact that (1) the user language may not be the same as the retrieved document(s) language, and (2) the documents may be written in multiple languages. \citet{liu2025xrag} provide a benchmark containing questions that require reasoning. They show that the language difference between the user and document languages can cause issues such as answers in the wrong language. They also show that documents of different languages make cross document reasoning more challenging. 
\citet{ranaldi2025multilingual} show a simple yet effective method for overcoming both issues; they use a translation service to translate the query and documents to english, then translate the answer. In contrast, \citet{wu2024not} show (on different benchmarks) that this translation-based method breaks down when using lower-quality translation systems, such as medium-scale LLMs. \citet{chirkova2024retrieval} provide practical solutions to the issue of a different user and document language; they highlight comments that when added to the system prompt, result in improved performance. \citet{qi_consistency_2025} focus on generation in cross-lingual RAG settings, addressing the influence of retrieved passages both when they are relevant, regardless of their language, or when distracting passages in different languages are provided in the context.

Several studies have examined bias in both retrieval and generation, namely the preference for high-resource languages like English over low-resource ones such as Arabic. \citet{wu2024not} evaluate end-to-end RAG performance across multiple LLMs and show that high-resource languages consistently outperform low-resource ones in both monolingual and cross-lingual settings. They also find that, when relevant documents exist in multiple languages, English passages are more likely to be selected. \citet{sharma2025faux} manually constructs a small benchmark over a synthetic corpus to avoid the influence of the parametric memory, and observe a consistent bias favoring the user query language in both stages. \citet{park_investigating_2025} analyze language preferences in both retrieval and generation, highlighting a strong bias toward high-resource languages, especially when the query and document languages match. English is noted as an exception, often outperforming even monolingual configurations -- an effect attributed to English dominance in pretraining data. 

Most prior work on multilingual RAG, including those cited here, relies on Wikipedia-based datasets and derived benchmarks such as MKQA \cite{longpre2021mkqa}, XOR-QA \cite{asai_xor_2021}, and MLQA \cite{lewis_mlqa_2020}. However, Wikipedia introduces several inherent properties: it is significantly richer in English content, has been typically used during the pretraining of both retrievers and generators, and much of its factual knowledge is encoded in the model’s parametric memory. All these factors impact cross-lingual behavior, and in particular, the behavior and influence of retrieval. \citet{chirkova2024retrieval}, while focusing on benchmarks derived from Wikipedia, explicitly acknowledge that retrieval performance in multilingual specialized domains remains under-explored. 

Thus, our work addresses a gap that has received limited attention by focusing on the retrieval component in a domain-specific, bilingual corporate setting involving a high- and low-resource language pair (English-Arabic). It uses clean multilingual corpora with well-aligned content across both languages, which are unlikely to have been seen during pretraining and represent realistic and practical RAG use cases.

\section{Evaluation Pipeline}
We use a cross-lingual basic RAG setup focused on English and Arabic. Given a query in either language, its goal is to generate an answer in the same language. The corpus includes documents in both languages, and each query is associated with a ground-truth answer found in one language only. The other language may contain partial or no relevant information. 

Our RAG pipeline consists of the standard components: retrieval, re-ranking, and answer generation. Retrieval is performed using dense vector search over a bilingual corpus\footnote{We split documents into passages, using LlamaIndex’s SentenceSplitter into passage of up to 100 tokens with no overlap. To preserve context, each passage retained the original document title, which corresponds to the law in the Legal benchmark and to the country in the Travel benchmark.}. We experiment with the multilingual embedding models {\em BAAI BGE-M3}\footnote{\url{https://huggingface.co/BAAI/bge-m3}} 
(referred to as \BGE from now on) and {\em Multilingual-E5-Large}\footnote{\url{https://huggingface.co/intfloat/multilingual-e5-large}} (referred to as \Efive), both of dimension $1024$, along with the {\em BGE-v2-M3}\footnote{\url{https://huggingface.co/BAAI/bge-reranker-v2-m3}} re-ranker. 
These models were chosen for their recency, popularity, and status as top-performing open-source retrievers and re-rankers \cite{li2023making,chen2024bge,wang2024multilingual,enevoldsen2025mmteb}.

For answer generation, we use Qwen-2.5-14B-Instruct\footnote{\url{https://huggingface.co/Qwen/Qwen2.5-14B-Instruct}}, a generative language model with strong multilingual capabilities, and part of the Qwen-2 family \citep{qwen2024}. During inference, the 20 most relevant passages are retrieved for a given question, then re-ranked based on their relevance and utility for answer generation. The top-5 ranked passages are used to augment the prompt provided to the LLM for answer generation (using Prompt~\ref{prompt:rag}).

\subsection{Metrics}\label{sec:metrics}

An effective RAG system requires success at three stages: retrieving a relevant passage, preserving it through re-ranking, and leveraging it in generation to produce an accurate answer.  We analyze the overall end-to-end performance, as well as each component in isolation: retrieval, re-ranking, and generation.

The end-to-end performance and the generation component are evaluated using an answer quality metric, which we refer to as accuracy, based on a semantic equivalence to ground-truth answers provided by our benchmarks (see Section~\ref{subsec:benchmarks}).
Specifically, we adopt an LLM-as-a-judge approach to assess correctness, using Claude 3.5 Sonnet to determine whether a generated answer matches the ground-truth reference (see Prompt~\ref{prompt:semantic}), following recent work by \citet{zheng-judging-llm}. Although LLM-based judgments have faced critique, particularly for relevance assessment \cite{soboroff2024}, prior studies have shown a high correlation with human evaluations in QA contexts. 
Moreover, the common alternative of strict lexical match is even less reliable in a multilingual setting, as discussed for example in \citet{qi_consistency_2025}, making a semantic measure more appropriate.

To further support this choice, we validated the metric through human evaluation with native speakers of the tested languages, confirming over 95\% agreement between human and automated ratings for both English and Arabic (see  Appendix~\ref{app:evaluation:accuracy} for more details). Given our focus on semantic similarity with respect to the ground truth, we find LLM-as-a-judge to be a practical and reliable measure.

For evaluating the retrieval component, we measure whether the ground-truth answer can be inferred from each retrieved passage. We obtain these relevance judgments using Claude 3.5 Sonnet with Prompt~\ref{prompt:retrieval}. Based on these relevance labels, we report Hits@20, indicating whether a relevant passage appears among the top $20$ retrieved results. For reranking, we apply the same procedure and report Hits@5 to measure whether relevant passages appear among the top results of the reranked list. Measuring the presence of relevant passages among the top results is particularly important in a RAG setting, as it reflects whether downstream components have access to the required evidence. The validity of these metrics is supported by their correlation with downstream accuracy, as detailed in Appendix~\ref{app:evaluation:hit}. Finally, to demonstrate that the Hit@20 results are consistent with other common metrics, Appendix~\ref{app:ndcg-mrr} also reports the NDCG and MRR corresponding to the results presented in this paper.

\subsection{Our Benchmarks}\label{subsec:benchmarks}
We focus on a corporate setting and construct two benchmarks, each based on a separate corpus. Both benchmarks are derived from public websites that contain parallel content in English and Arabic. The first benchmark, \textit{Legal}, is based on the UAE Legislation website\footnote{\url{https://uaelegislation.gov.ae/}}, which contains 390 laws, with each law described in separate documents in English and Arabic. The second benchmark, \textit{Travel}, is based on the UAE Ministry of Foreign Affairs website\footnote{\url{https://www.mofa.gov.ae/ar-ae/travel-updates}}, which offers travel-related information for multiple countries, such as visa requirements and embassy contacts. For each country, the information is presented in two parallel documents, one per language.

Having each document available in both languages is essential for our experimental design. In order to build a corpus for each of these two use cases, we assign a \textit{document language} to each document uniformly at random during corpus construction, ensuring that every document appears in exactly one language within the corpus. 
The resulting Legal corpus includes roughly 1.5M  words, while the Travel corpus contains around 150K words.
After building and indexing this bilingual corpus, we proceeded to create the benchmark. We used DataMorgana \cite{filice2025datamorgana}, a synthetic question–answer generation tool, to create query–answer pairs per document, ensuring that each question could be answered using that document alone. The language of each query–answer pair (the \textit{user language}) is also selected uniformly at random and independently of the document language, resulting in a benchmark that supports systematic evaluation across all language combinations, and allows to identify the source of bias. 
The final benchmarks include around 1.3K question–answer pairs for Legal and 2K for Travel.
Details of the DataMorgana configuration we used to generate our benchmarks, along with basic statistics, are provided in Appendix~\ref{app:datamorgana}\footnote{Our benchmarks and corpora are available at: \url{https://github.com/chenamiraz/cross-lingual-cost}. In the Legal index, the law id serves as the document id, while in the Travel index, the country name is used as the title. }.

\section{Experiments}\label{sec:experiments}
We present four experiments, each structured with a description, results, and key conclusions. The first experiment demonstrates that retrieval is a major bottleneck in our bilingual setting.
The second reveals performance gaps between same-language and cross-lingual cases, with substantially worse results when the user and document languages differ. The third attributes this performance drop to the retriever’s need to rank documents in both languages against the query simultaneously.
Finally, the fourth proposes and evaluates mitigation strategies to address this issue.

\subsection{Retrieval is a Critical Bottleneck}
Table~\ref{tab:components} presents the results of our first experiment, using the metrics described in Section~\ref{sec:metrics}. We first measured accuracy without retrieval augmentation for each benchmark. Then, for each of our two embedding models, we evaluated the performance of each system component as well as the overall end-to-end performance.

Specifically, we report Hits@20 for the retrieval phase. For 
reranking, we report Hits@5 only on examples where retrieval achieved Hits@20 equal to 1, meaning a passage with the answer was passed to the reranker. For generation, we report answer accuracy only on examples where reranking achieved Hits@5 equal to 1, namely where a passage containing the answer was included in the prompt. This analysis helps identify how each phase contributes to the overall end-to-end accuracy. 

\begin{table*}[ht]
\centering
\resizebox{\linewidth}{!}{%
\begin{tabular}{cc|c|ccc|c}
\toprule
{\bf Benchmark} & {\bf No-RAG} & {\bf Embedder}  & {\bf Retrieval} & {\bf Reranking} & {\bf Generation} & {\bf End-to-End} \\
\midrule
\multirow{2}{*}{Legal} &\multirow{2}{*}{27±3\%} 
&\BGE  & 81±2\% & 88±2\% & 78±3\% & 58±3\% \\
&&\Efive & 66±3\% & 87±2\% & 78±3\% & 48±3\% \\
\hline
\multirow{2}{*}{Travel} &\multirow{2}{*}{37±3\%}
&\BGE & 89±1\% & 97±1\% & 87±2\% & 79±2\% \\
&&\Efive & 76±2\% & 97±1\% & 85±2\% & 67±2\% \\
\bottomrule
\end{tabular}
}
\caption{{\bf No-RAG baseline and RAG component-wise and end-to-end performance.} For each benchmark, we first report the baseline answer accuracy using only the user question without retrieval augmentation, referred to as No-RAG. Then, for each embedding model, we report the retriever Hit@20, the reranker Hit@5 conditioned on successful retrievals, the generation answer accuracy conditioned on successful rerankings, and the overall end-to-end answer accuracy. Each value is presented with its 95\% confidence interval.}
\label{tab:components}
\end{table*}

The Legal benchmark represents a domain-specific setting, where questions involve niche topics, so the LLM cannot rely on its parametric memory alone to answer them, as shown by the low accuracy achieved without RAG. This is further confirmed by comparing the end-to-end score in Table \ref{tab:components} with the product of retrieval score, reranking score conditioned on successful retrieval, and generation score conditioned on successful reranking. These values are nearly identical, indicating that the generation phase cannot compensate for failures earlier in the pipeline. The table shows similar results for the Travel benchmark, although the overall accuracy for this case is slightly higher than the product of the component-level conditional scores. This is likely because the Travel corpus includes less specialized knowledge, making it better represented in the LLM’s parametric memory, as also reflected by the performance gap without retrieval. 

Looking more closely at the individual components, the reranker performs the best of the three. For both benchmarks with the \BGE embedder, the probability of retrieval failure is comparable to that of generation. With the \Efive embedder, the retrieval gap is even larger than the generation gap, showing a 12\% difference on the Legal benchmark and 9\% on Travel. 
Moreover, for each benchmark, reranking and generation performance are stable across embedders. However, changing retrievers has a substantial effect on end-to-end accuracy.
These results, taken together, highlight that the retriever is a critical bottleneck and motivate us to focus our efforts on it.

\subsection{Cross-Lingual Combinations are the Most Challenging}
\begin{table*}[ht]
  \centering
  \begin{minipage}{0.48\textwidth}
\centering
\resizebox{\linewidth}{!}{%
\begin{tabular}{c|cc|cc}
\toprule
\textbf{Benchmark} & \makecell{\textbf{User} \\ \textbf{Lang.}} & \makecell{\textbf{Doc} \\ \textbf{Lang.}} & \makecell{\textbf{Retrieval}\\ \textbf{Hit@20}} & \makecell{\textbf{End-to-End}\\ \textbf{Accuracy}}  \\
\hline
\multirow{5}{*}{Legal} 
  & Arabic  & Arabic  & 92±3\% & 68±5\% \\
  & Arabic  & English & 90±3\% & 67±5\% \\
  & English & Arabic  & 56±5\% & 31±5\% \\
  & English & English & 86±4\% & 68±5\% \\
  \cdashline{2-5}
  & \multicolumn{2}{c|}{Same-lang.} & 89±2\% & 68±4\% \\
  & \multicolumn{2}{c|}{Cross-lang.} & 73±3\% & 49±4\% \\
\hline
\multirow{5}{*}{Travel} 
  & Arabic  & Arabic  & 93±2\% & 85±3\% \\
  & Arabic  & English & 91±3\% & 78±4\% \\
  & English & Arabic  & 80±4\% & 70±4\% \\
  & English & English & 94±2\% & 84±3\% \\
  \cdashline{2-5}
  & \multicolumn{2}{c|}{Same-lang.} & 93±2\% & 84±2\% \\
  & \multicolumn{2}{c|}{Cross-lang.} & 86±2\% & 74±3\% \\
\bottomrule
\end{tabular}
}
\subcaption{{ \BGE embedder} }
\label{tab:per-language-bge}
  \end{minipage}%
  \hfill
  \begin{minipage}{0.48\textwidth}
    \centering
\resizebox{\linewidth}{!}{%
\begin{tabular}{c|cc|cc}
\toprule
\textbf{Benchmark} & \makecell{\textbf{User} \\ \textbf{Lang.}} & \makecell{\textbf{Doc} \\ \textbf{Lang.}} & \makecell{\textbf{Retrieval}\\ \textbf{Hit@20}} & \makecell{\textbf{End-to-End}\\ \textbf{Accuracy}}  \\
\hline
\multirow{5}{*}{Legal} 

  & Arabic  & Arabic  & 87±4\% & 67±5\% \\
  & Arabic  & English & 51±5\% & 37±5\% \\
  & English & Arabic  & 41±5\% & 22±4\% \\
  & English & English & 88±4\% & 70±5\% \\
  \cdashline{2-5}
  & \multicolumn{2}{c|}{Same-lang.} & 88±3\% & 69±4\% \\
  & \multicolumn{2}{c|}{Cross-lang.} & 46±4\% & 29±3\% \\
  
\hline
\multirow{5}{*}{Travel} 
  & Arabic  & Arabic  & 90±3\% & 86±3\% \\
  & Arabic  & English & 54±4\% & 37±4\% \\
  & English & Arabic  & 64±4\% & 60±4\% \\
  & English & English & 95±2\% & 85±3\% \\
  \cdashline{2-5}
  & \multicolumn{2}{c|}{Same-lang.} & 92±2\% & 86±2\% \\
  & \multicolumn{2}{c|}{Cross-lang.} &  59±3\% & 49±3\% \\

\bottomrule
\end{tabular}
}
\subcaption{{ \Efive embedder} 
}
\label{tab:per-language-e5}
  \end{minipage}
\caption{{\bf Performance across language combinations.} 
Results are presented for each embedder, benchmark and for each of the four possible user–document language combinations. In addition, we report same-language and cross-language scores, defined as the mean scores over combinations where the user and document languages match or differ, respectively. Each value is presented with its 95\% confidence interval.
}
\label{tab:per-language}
\end{table*}

Next, we compare the retrieval and end-to-end performances on each of the four user-document language combinations. The results for the \BGE and \Efive embedders are presented in Tables~\ref{tab:per-language-bge} and~\ref{tab:per-language-e5}, respectively.

The tables reveal that cross-lingual scenarios, where the user query and the supporting document are in different languages, consistently underperform compared to same-language settings in both retrieval and end-to-end performance. 
For the \BGE embedder, a substantial decline in retrieval performance is observed only when the user language is English and the document language is Arabic, with drops of 33\% in the Legal benchmark and 13\% in Travel compared to the same-language configuration. A similar pattern appears in the final accuracy, with decreases of 37\% and 14\%, respectively. Notably, the reverse cross-lingual setting does not exhibit any statistically significant degradation for \BGE.

In contrast, the \Efive embedding exhibits an even larger performance drop across both cross-lingual settings. Specifically, retrieval Hit@20 decreases by 42\% on the Legal benchmark and by 33\% on Travel, compared to their same-language counterparts. These retrieval declines also propagate to the end-to-end accuracy, resulting in drops of 40\% for Legal and 37\% for Travel.

In what follows we dive deeper to discover the cause behind this gap.

\subsection{The Source of the Cross-Lingual Failure}
Notice that in our current setup, referred to from now on as the \direct setting, we face two key challenges due to multilinguality. Firstly, ``query-document language mismatch'' requires the retriever to rank documents in one language in response to queries in another. Secondly, ``document-document language mismatch'' necessitates ranking documents across various languages without favoring high-resource languages or the user's language.

To determine which of these challenges is primarily responsible for the observed failures, we conducted the following experiment. We modified the retriever from the \direct setting to search only within the correct language. Specifically, for query corresponding to a (ground truth) document language X, the \oracle retriever returns the top results exclusively in language X, completely excluding language Y. Hence, the \oracle retriever has the "query-document language mismatch" challenge but completely avoids the "document-document language mismatch" challenge.

\begin{figure*}[ht]
  \centering
  \begin{tabular}{cc}
    \begin{subfigure}[t]{0.45\textwidth}
      \centering
      \caption{Legal benchmark – \BGE embedder}
      \includegraphics[width=\linewidth]{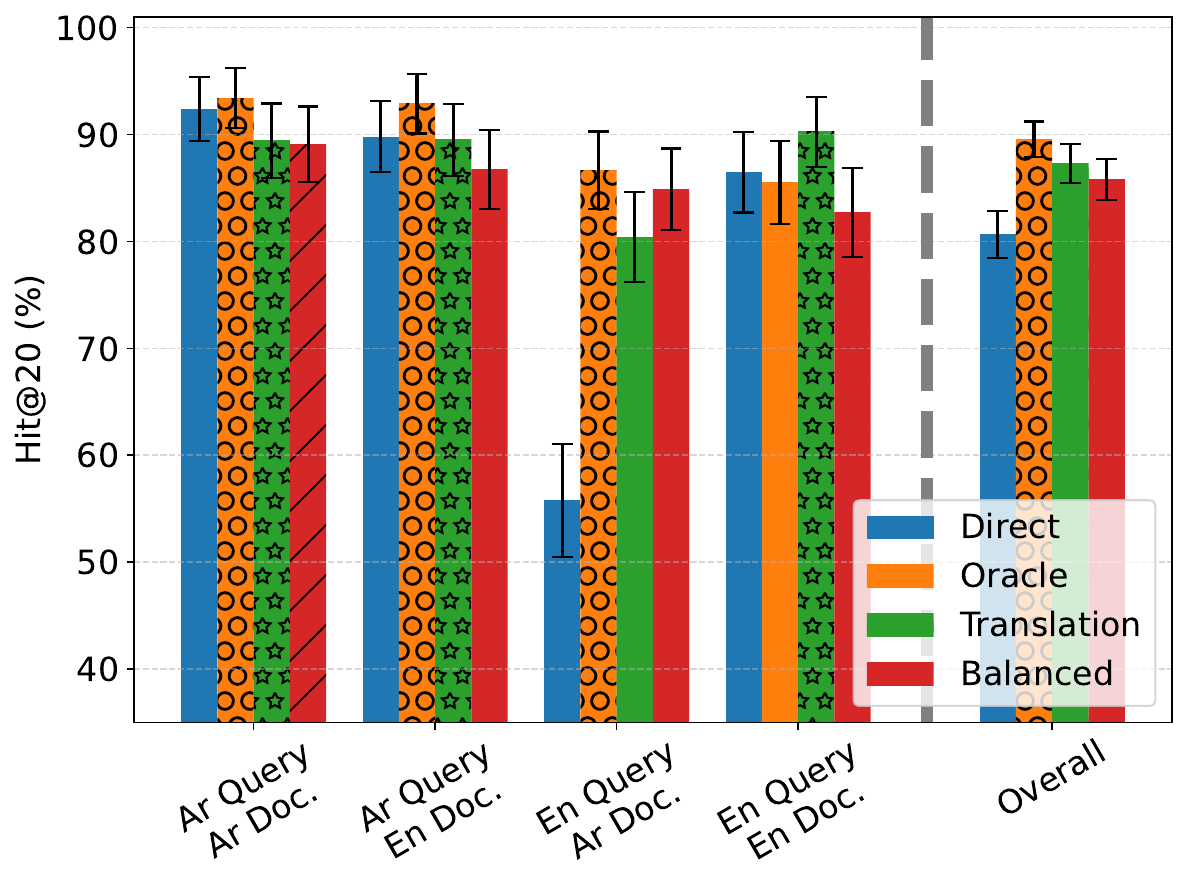}
      \label{fig:hit20:legal:bge}
    \end{subfigure} &
    \begin{subfigure}[t]{0.45\textwidth}
      \centering
      \caption{Legal benchmark – \Efive embedder}
      \includegraphics[width=\linewidth]{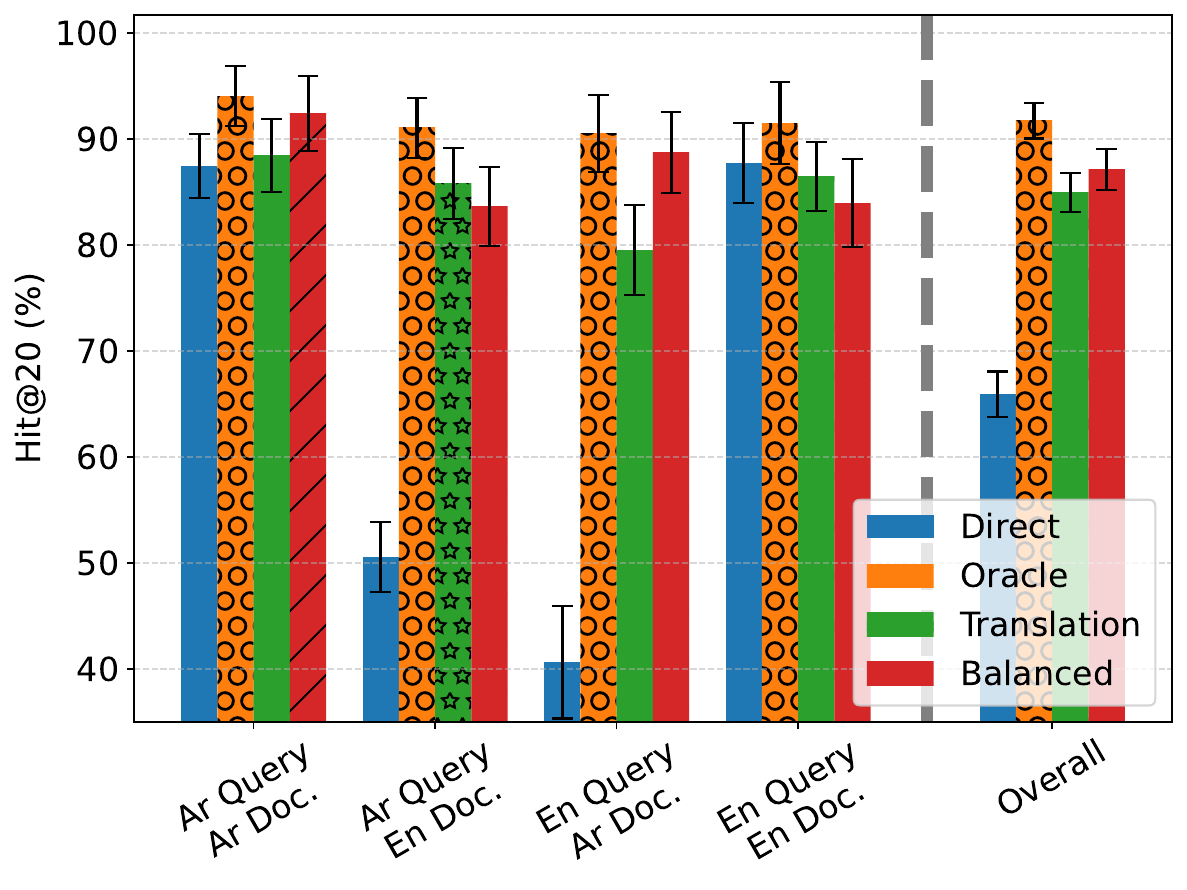}
      \label{fig:hit20:legal:e5}
    \end{subfigure} \\

    \begin{subfigure}[t]{0.45\textwidth}
      \centering
      \caption{Travel benchmark – \BGE embedder}
      \includegraphics[width=\linewidth]{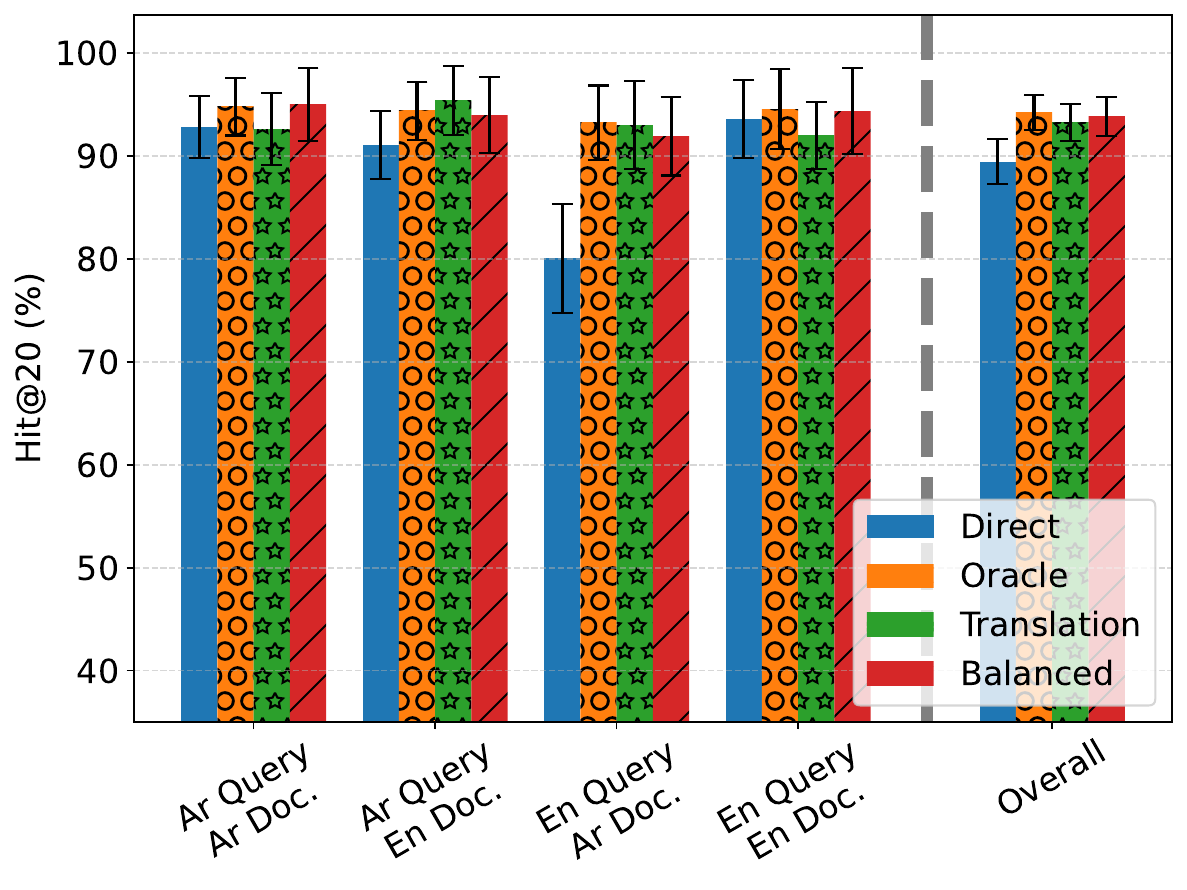}
      \label{fig:hit20:mofa:bge}
    \end{subfigure} &
    \begin{subfigure}[t]{0.45\textwidth}
      \centering
      \caption{Travel benchmark – \Efive embedder}
      \includegraphics[width=\linewidth]{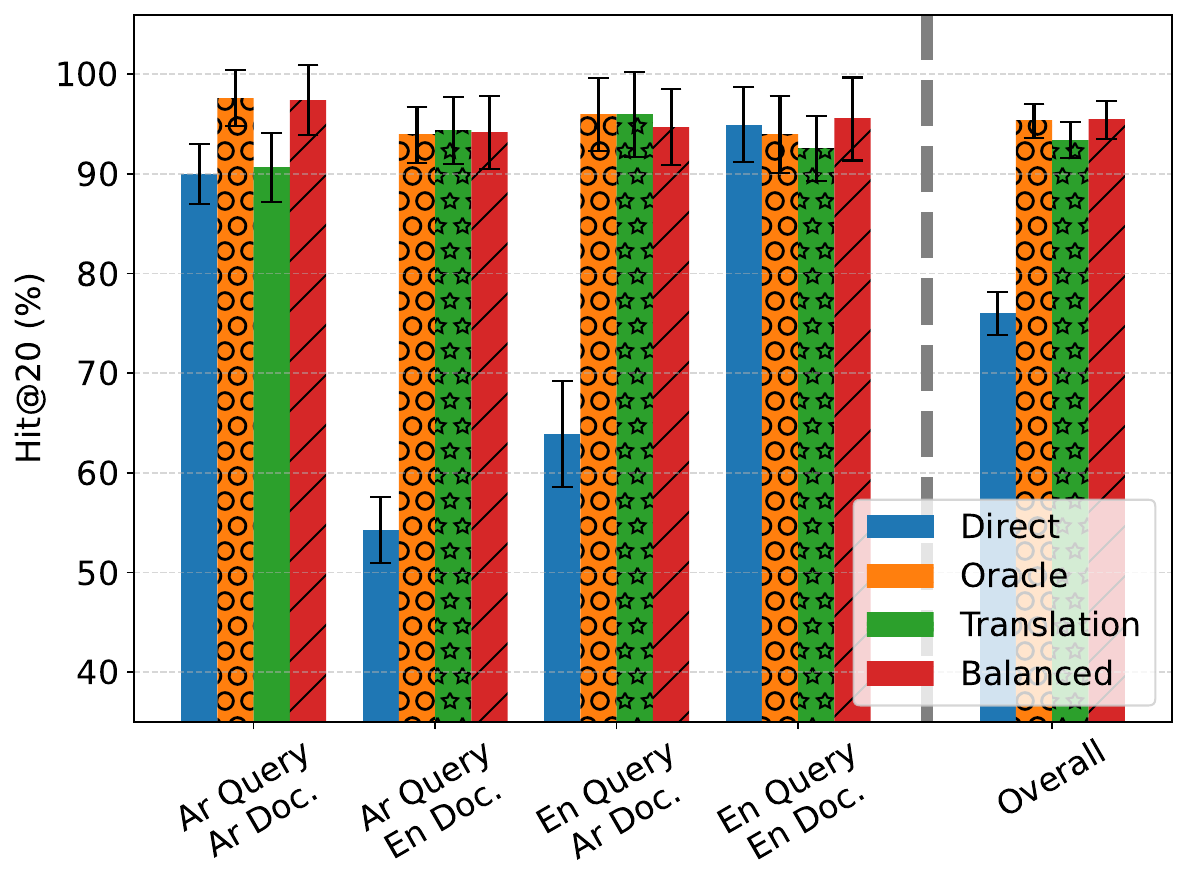}
      \label{fig:hit20:mofa:e5}
    \end{subfigure}
  \end{tabular}

  \caption{{\bf Retrieval Hit@20 scores across benchmarks and embedders.} Each figure corresponds to a specific combination of benchmark and embedding.
Bars represent retrieval Hit@20 scores in percentages, with 95\% confidence intervals shown as black error lines.
Different retrieval policies are distinguished by color and texture.
Results are grouped by benchmark segments defined by the user-document language combination, as well as the overall benchmark retrieval  performance.}
  \label{fig:hit20:all}
\end{figure*}

We stress that the oracle is used only for analysis purposes, since in practice we do not have access to the document language ahead of time. 
The first two bars in each subfigure of Figure~\ref{fig:hit20:all} present the Hit@20 performance of the \direct and \oracle retrievers, broken down by query-document language combinations, as well as overall indicating the performance over the entire benchmark.

We observe two clear phenomena. First, the \oracle retriever achieves nearly identical performance across all query and document language pairs, suggesting there are essentially no failures related to the query-document language mismatch challenge. 
In contrast, the gap between the \direct and \oracle retrievers can be substantial in many cross-lingual cases. This indicates that the main source of failure lies in the
document-document language mismatch challenge, namely the retriever’s ability to rank documents across languages. 

The results suggests that while semantic similarity is well captured within a single language, the retrievers struggle in cross-lingual settings. For instance, \BGE appears to favor English passages when the user query is in English, while \Efive may exhibit a tendency to prefer passages in the same language as the user query.

\subsection{Mitigating Cross-lingual Failings}
These results raise an important question: can multilingual retrievers be used reliably on mixed-language corpora without further tuning?
To address this question, we consider two retrieval baselines.  The first, denoted \translation, translates each query into the other language using the Google Translate API and performs retrieval twice (once per language, since the document language is not known a priori). The two ranked lists are then merged, and the top 20 results are selected according to the retrieval score (i.e., the inner product of the query and document embeddings).
The second method, \balanced, enforces equal selection across languages by retrieving 10 passages in Arabic and 10 in English. 

We evaluate these approaches under the same experimental setup described earlier. The last two bars in each subfigure of Figure~\ref{fig:hit20:all} present the corresponding results. While the \oracle retriever is not feasible in practice, it serves as an upper bound for what the \translation and \balanced approaches could achieve.
In practice, both the \translation and \balanced retrievers show no statistically significant loss relative to the \direct retriever in same-language cases, while providing substantial improvements in cross-lingual cases. Notably, those retrievers yields more consistent results across the different combinations of user and document languages, unlike the \direct setting, which favors the same-language combinations at the expense of cross-language ones. Moreover, this strategy leads to a considerable improvement in overall retrieval accuracy across benchmarks and embedders, with consistent gains of around 4-6\% for \BGE and approximately 20\% for \Efive.

No statistically significant difference is observed between the performance of the \translation retriever and the \balanced retriever. However, they differ in latency and cost: while \translation requires an expensive and time-consuming call to a translation service, \balanced incurs no additional cost beyond retrieving documents from the index. One might also assume that \balanced requires prior knowledge of the proportion of ground-truth documents in each language. Yet, as shown in Appendix~\ref{app:imbalanced}, the performance of \balanced remains stable even when this proportion varies, suggesting that \balanced may offer a more practical alternative in such scenarios.

\section{Conclusions}\label{sec:conclusion}
This work highlights retrieval as a critical bottleneck in multilingual RAG systems applied to domain-specific corpora. While prior studies have identified and focused on generation as the main limitation in cross-lingual RAG, their conclusions are primarily based on Wikipedia-derived benchmarks. Since multilingual retrievers such as BGE-M3 and multilingual-E5-large are trained on similar open-domain data, they exhibit strong performance in those settings. In contrast, our domain-specific benchmarks expose substantial retrieval weaknesses that remain obscured in such evaluations, underscoring the need to revisit cross-lingual retrieval in practical, real-world RAG scenarios.

Our analysis shows that performance degrades most in cross-lingual settings where the user and document languages differ, with drops that can exceed 40\% compared to same-language configurations. Using an oracle retriever restricted to the correct language, we isolate the primary source of failure as the retriever’s difficulty in ranking documents across languages. That is, while the retriever performs well within a single language, it struggles when comparing passages across languages, often favoring those in the query’s language. We further observe that different embedders exhibit weaknesses in different cross-lingual settings. This highlights the potential to improve training by explicitly targeting cross-lingual robustness and narrowing the gap with same-language performance.

Lastly, we show that simple mitigations, such as retrieving a balanced number of documents per language or translating the query, can substantially improve cross-lingual performance and even enhance overall results. This finding highlights meaningful opportunities for reducing multilingual retrieval biases, particularly in real-world applications. However, applying such approaches in practical settings with non-uniform language distributions or more than two languages remains an open challenge and warrants further investigation.

\paragraph*{Acknowledgments}
We thank our colleagues at AI71, and in particular Abdelrahman Ibrahim, Amr Ali Abugreedah, Mohamad Salah, Saleem Hamo, Kirollos Sorour, Imran Moqbel, and Anas AlHelali, whose native proficiency in Arabic was instrumental in the annotation process used to validate the evaluation metrics in our pipeline, as well as Michal Caspi for co-leading this effort. 
\bibliography{bib}

\appendix

\section{Appendix}\label{sec:appendix}

\subsection{Metric Evaluation}\label{app:evaluation}
\subsubsection{Answer Accuracy}\label{app:evaluation:accuracy}
To validate our answer accuracy metric across languages, we performed the following procedure. First, we used samples from English and Arabic editions of Wikipedia to construct two benchmarks of 100 examples each, using DataMorgana \cite{filice2025datamorgana}. We then applied a standard RAG pipeline to generate answers using Falcon-3-10B.  The generated answers were then compared to reference answers using our LLM-as-a-judge-based accuracy metric, as described in the main text.  This setting was intentionally selected to produce a mix of correct and incorrect answers, ensuring a meaningful evaluation of the metric.

Independently, human annotators who are native speakers of the respective languages were asked to assess the similarity between the generated and reference answers. The annotators were asked to label each pair as matching or not matching and to mark them as debatable or non-debatable. Of the non-debatable items (80\% in both languages) the agreement rate was 95\% for English and 98\% for Arabic. The overall agreement rates are 82\% for English and 85\% for Arabic, which means that almost all disagreements were for cases marked as debatable. Therefore, the annotations corroborate the validity of the automated accuracy metric.

\subsubsection{Retrieval Hit@20}\label{app:evaluation:hit}

Now that we trusted our LLM-based accuracy metric, we moved to validating whether our Hits@20 metric, which also uses LLM judgments, effectively captures success in the retrieval step. Toward this goal, we analyzed the downstream accuracy as a function of the Hit@20 score. This analysis focused on the Legal benchmark, where the no-RAG accuracy is relatively low (27\%), making it easier to observe the impact of retrieval quality. 
Table~\ref{tab:metrics} reports these results for the \BGE and \Efive embedders.

\begin{table}[t]
\centering
\begin{tabular}{lcc}
\toprule
\multicolumn{1}{c}{} & \multicolumn{2}{c}{\textbf{End-to-End Accuracy}} \\
\cmidrule(lr){2-3}
\textbf{Retrieval Hit@20} & \multicolumn{1}{c}{\BGE} & \multicolumn{1}{c}{\Efive} \\
\midrule
0 & 10±3\% & 9±2\% \\
1 & 79±2\% & 79±3\% \\
Overall & 60±2\% & 50±2\% \\
\bottomrule
\end{tabular}
\caption{{\bf End-to-End Accuracy as Function of Our LLM-based Hit@20.} Each cell shows the average accuracy along with its 95\% confidence interval. Columns correspond to retrieval embedders; rows indicate evaluation segments: instances with Hit@20 = 0, Hit@20 = 1, and overall accuracy.}
\label{tab:metrics}
\end{table}

As shown in Table~\ref{tab:metrics}, downstream accuracy was indeed low when the Hits@20 metric indicates failure, confirming that our LLM-based Hits@20 reliably identifies cases where retrieval has failed. Specifically, accuracy dropped to approximately 9\% when no relevant passage was identified by the metric, which is considerably lower than the 27\% accuracy observed without retrieval augmentation. Furthermore, we observed consistent patterns across retrievers: although the \BGE retriever differed markedly in overall quality from the \Efive retriever, their downstream accuracy as a function of retrieval quality showed only minor differences, likely attributable to statistical noise.
These findings validate our Hits@20 metric as a reliable measure of retrieval effectiveness, demonstrating that higher scores are strongly associated with improved downstream accuracy.

\subsection{Benchmark configuration and statistics}\label{app:datamorgana}
The following describes the configuration used to construct both the Legal and Travel benchmarks. In both cases, DataMorgana was configured in non-conversational mode, supporting single-turn question answering only.

DataMorgana allows the definition of multiple parallel question categorizations, each selected independently of the rest of the configuration, including other categories and the document language. The question categorizations were defined as follows:
\begin{itemize}
    \item \textbf{Language:} The user language was set to Arabic in 50\% of the cases and English in 50\%.
    
    \item \textbf{Formulation:} The question was phrased as:
    \begin{itemize}
        \item Concise natural language: 40\% of cases.
        \item Verbose natural language: 20\% of cases.
        \item Short search query: 25\% of cases.
        \item Long search query: 15\% of cases.
    \end{itemize}
    
    \item \textbf{Linguistic similarity:} In 50\% of the cases, the phrasing was similar to that found in the corpus, and in the remaining 50\%, it had a greater linguistic distance.
    
    \item \textbf{Question type:} Questions were evenly split between factoid (50\%) and open-ended (50\%).
    
    \item \textbf{User need:}
    \begin{itemize}
        \item For the Legal benchmark, 50\% of the questions simulated a user seeking specific legal advice, while the other 50\% simulated a user asking out of general curiosity.
        \item For the Travel benchmark, the user type was distributed as follows: UAE user in 20\% of the cases, Non-UAE user in an additional 30\%, and Undisclosed citizenship in the remaining 50\%.
    \end{itemize}
\end{itemize}
The benchmark was balanced after the DataMorgana filtering step to include 50\% questions grounded in Arabic documents and 50\% in English documents. Statistics for the final benchmark are presented in Table~\ref{tab:benchmarks}.
\begin{table}[h!]
\centering
\resizebox{\columnwidth}{!}{%
\begin{tabular}{l l l r}
\hline
Benchmark & \makecell{Query \\ language} & \makecell{Document \\ language} & Count \\
\hline
\multirow{4}{*}{Legal} 
  & English & English & 318 \\
  & English & Arabic  & 337 \\
  & Arabic  & English & 324 \\
  & Arabic  & Arabic  & 303 \\
\hline
\multirow{4}{*}{Travel} 
  & English & English & 513 \\
  & English & Arabic  & 471 \\
  & Arabic  & English & 479 \\
  & Arabic  & Arabic  & 460 \\
\hline
\end{tabular}%
}
\caption{Benchmark breakdown by query and document language.}
\label{tab:benchmarks}
\end{table}

\subsection{Prompts}
In this section, we provide all the prompts used in our experiments.
Prompt~\ref{prompt:rag} was used for answer generation. It is based on the guidelines proposed by \citet{chirkova2024retrieval} for prompting RAG systems in multilingual scenarios.
Prompt \ref{prompt:semantic} was used to evaluate the accuracy of the generated answer.
Prompt \ref{prompt:retrieval} was used to evaluate retrieval Hit@20 and reranking Hit@5.

\begin{promptbox}[prompt:rag]{RAG generation}
\paragraph{System.}
Answer the question based on the given passages below.

Elaborate when answering, and if applicable provide additional helpful information from the passages and only from the passages.
Do not refer to the passages, just state the information.

You MUST answer in the SAME LANGUAGE as the QUESTION LANGUAGE, regardless of the language of the passages. Answering in the same language as the user is asking their question is crucial to your success. If the question is in English, the answer must also be in English. If the question is in Arabic, the answer must also be in Arabic.

Write all named entities in the same language and same alphabet as the question language.

\paragraph{User.}
\# Passages:

passage 1:

<Passage 1>

passage 2:

<Passage 2>

passage 3:

<Passage 3>

...

\# Question: <Question>
\end{promptbox}

\begin{promptbox}[prompt:semantic]{Generated answer evaluation}
Based on the question and the golden answer, judge whether the predicted answer has the same meaning as the golden answer. Return your answer in the following format: <same\_meaning>True/False</same\_meaning>.

<question> ... </question>

<golden\_answer> ... </golden\_answer>

<predicted\_answer> ... </predicted\_answer>

\end{promptbox}

\begin{promptbox}[prompt:retrieval]{Retrieval evaluation}
You are given a **question**, a **ground truth answer**, and a list of **passages**.  
Your task is to return the **list of passage indices** that can directly answer the question **by containing the ground truth answer** (i.e., the passage includes a perfect match to the information expressed in the ground truth).  

Please follow these rules:

- A passage should be included only if it **clearly expresses or contains the ground truth answer**.

- Do **not include passages** that are only loosely related or provide background information.

- Your response **must be valid Python list syntax**, e.g., [3, 5, 9].

- Do **not add any explanation** outside the list.

---

**Question**:  
<Question>

**Ground Truth Answer**:  
<Answer>

**Passages**:  
Passage 1:
<Passage 1 content>

Passage 2:
<Passage 2 content>

Passage 3:
<Passage 3 content>

 ...
\end{promptbox}

\subsection{Additional Results with NDCG and MRR}\label{app:ndcg-mrr}
\begin{figure*}[ht]
  \centering
  \begin{tabular}{cc}
    \begin{subfigure}[t]{0.45\textwidth}
      \centering
      \caption{Legal benchmark – \BGE embedder}
      \includegraphics[width=\linewidth]{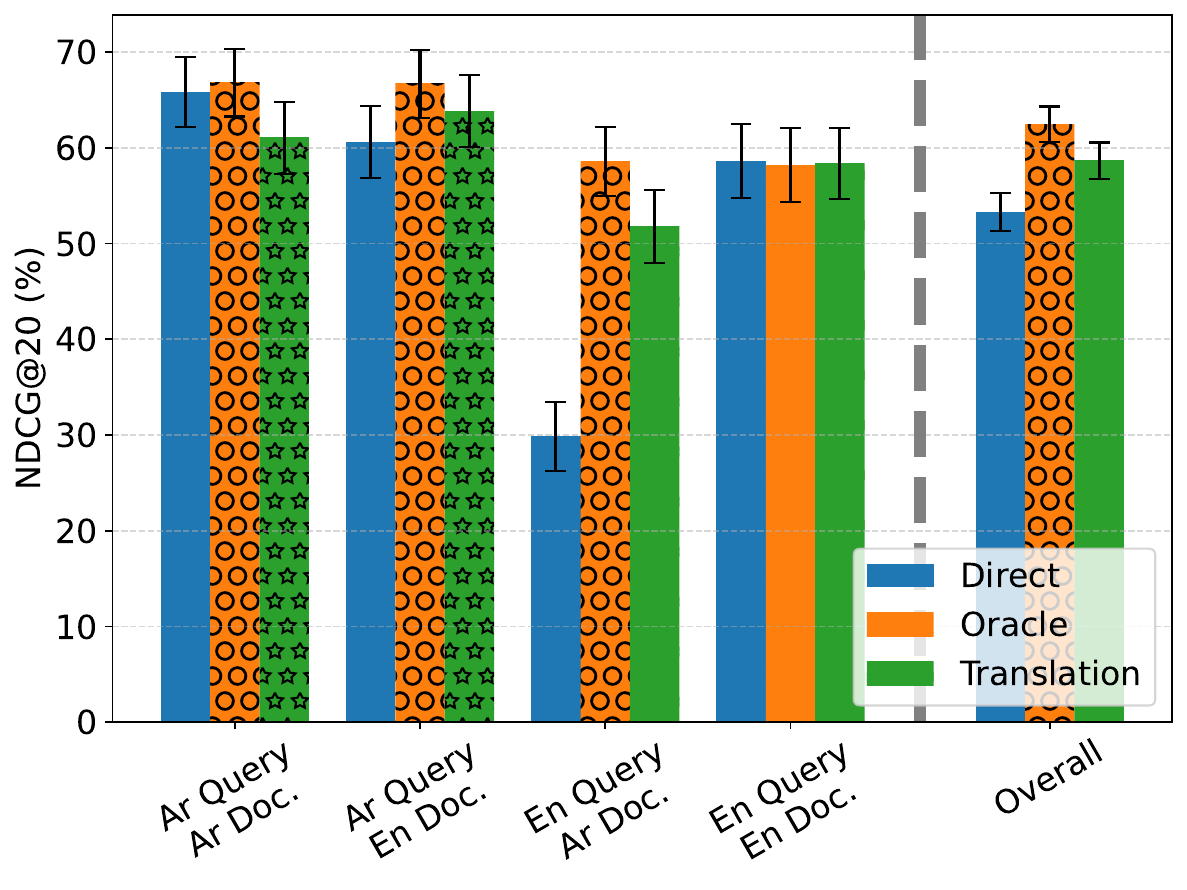}
      \label{fig:ndcg:legal:bge}
    \end{subfigure} &
    \begin{subfigure}[t]{0.45\textwidth}
      \centering
      \caption{Legal benchmark – \Efive embedder}
      \includegraphics[width=\linewidth]{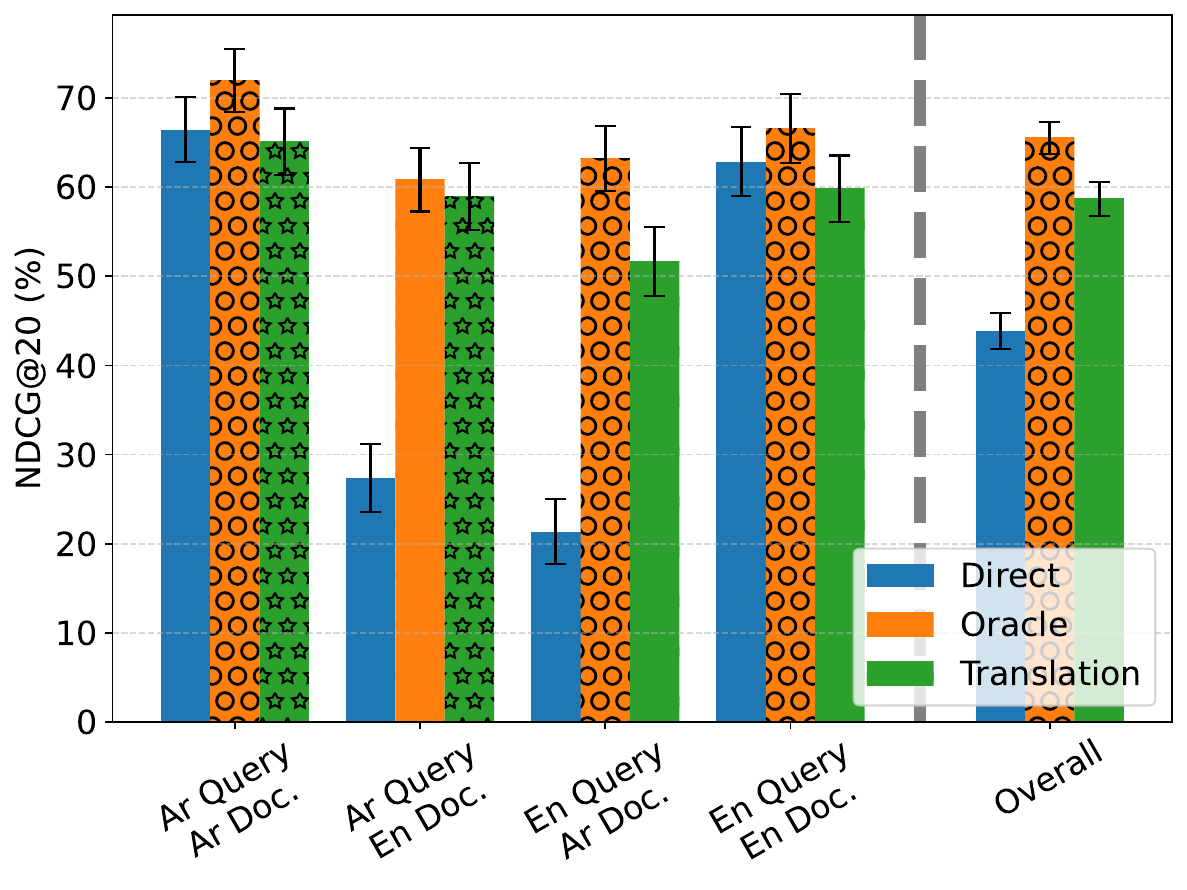}
      \label{fig:ndcg:legal:e5}
    \end{subfigure} \\

    \begin{subfigure}[t]{0.45\textwidth}
      \centering
      \caption{Travel benchmark – \BGE embedder}
      \includegraphics[width=\linewidth]{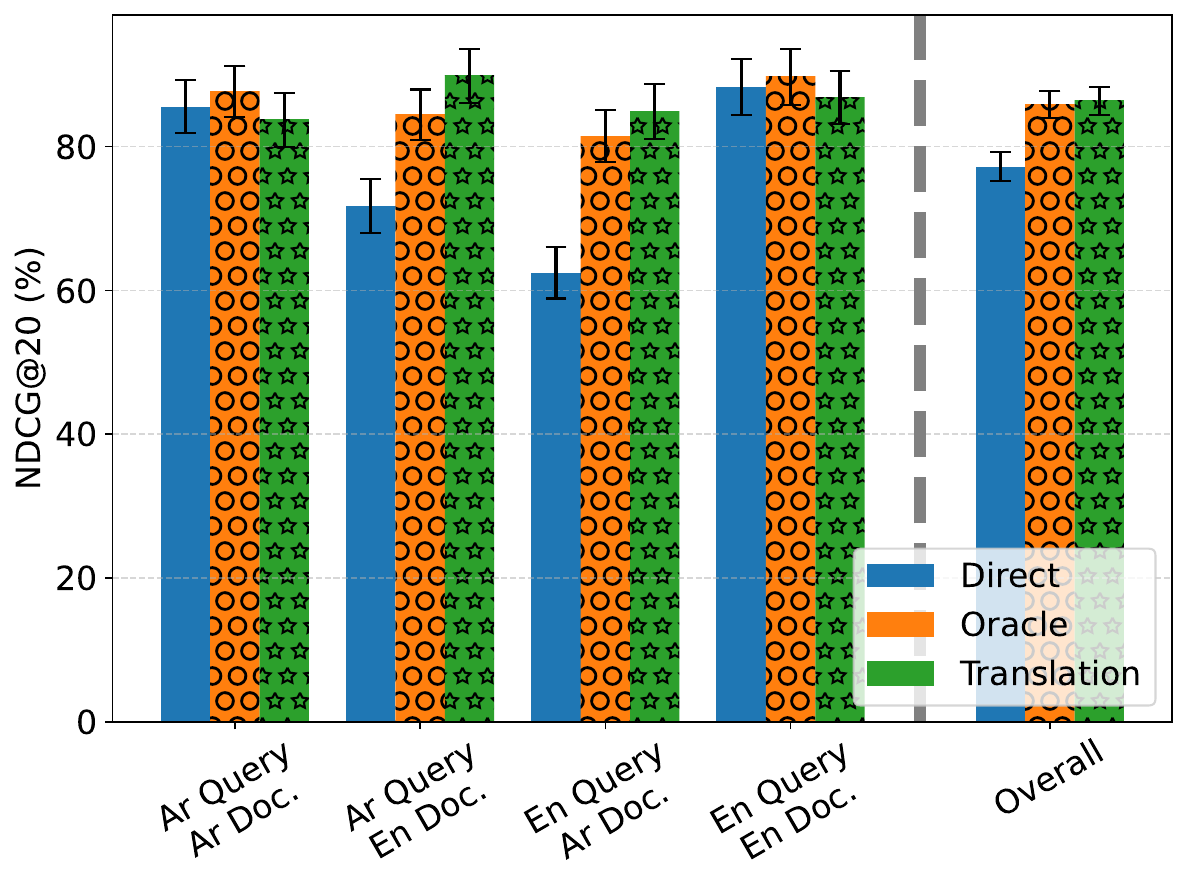}
      \label{fig:ndcg:mofa:bge}
    \end{subfigure} &
    \begin{subfigure}[t]{0.45\textwidth}
      \centering
      \caption{Travel benchmark – \Efive embedder}
      \includegraphics[width=\linewidth]{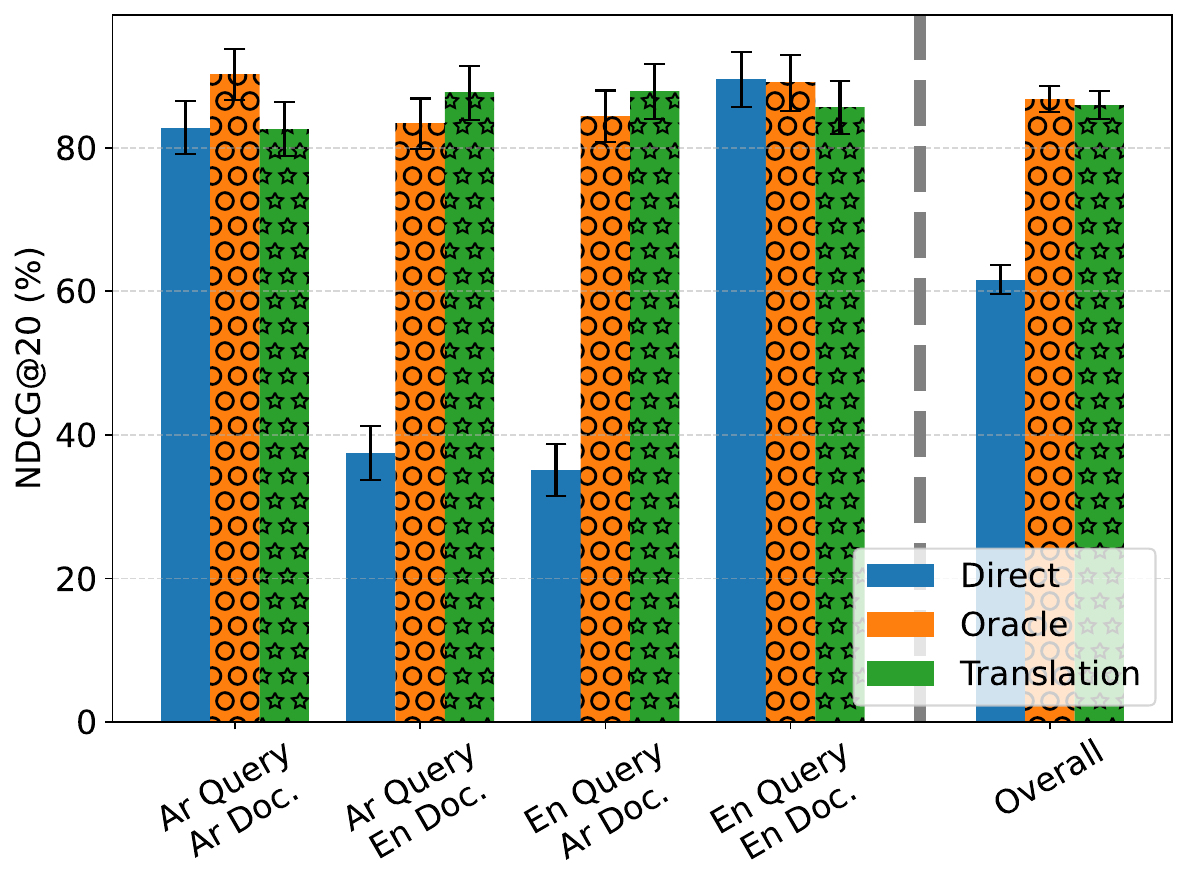}
      \label{fig:ndcg:mofa:e5}
    \end{subfigure}
  \end{tabular}

  \caption{{\bf Retrieval NDCG@20 scores across benchmarks and embedders.} Each figure corresponds to a specific combination of benchmark and embedding.
Bars represent retrieval NDCG@20 scores in percentages, with 95\% confidence intervals shown as black error lines.
Different retrieval policies are distinguished by color and texture.
Results are grouped by benchmark segments defined by the user-document language combination, as well as the overall benchmark retrieval performance.}
  \label{fig:ndcg:all}
\end{figure*}

\begin{figure*}[ht]
  \centering
  \begin{tabular}{cc}
    \begin{subfigure}[t]{0.45\textwidth}
      \centering
      \caption{Legal benchmark – \BGE embedder}
      \includegraphics[width=\linewidth]{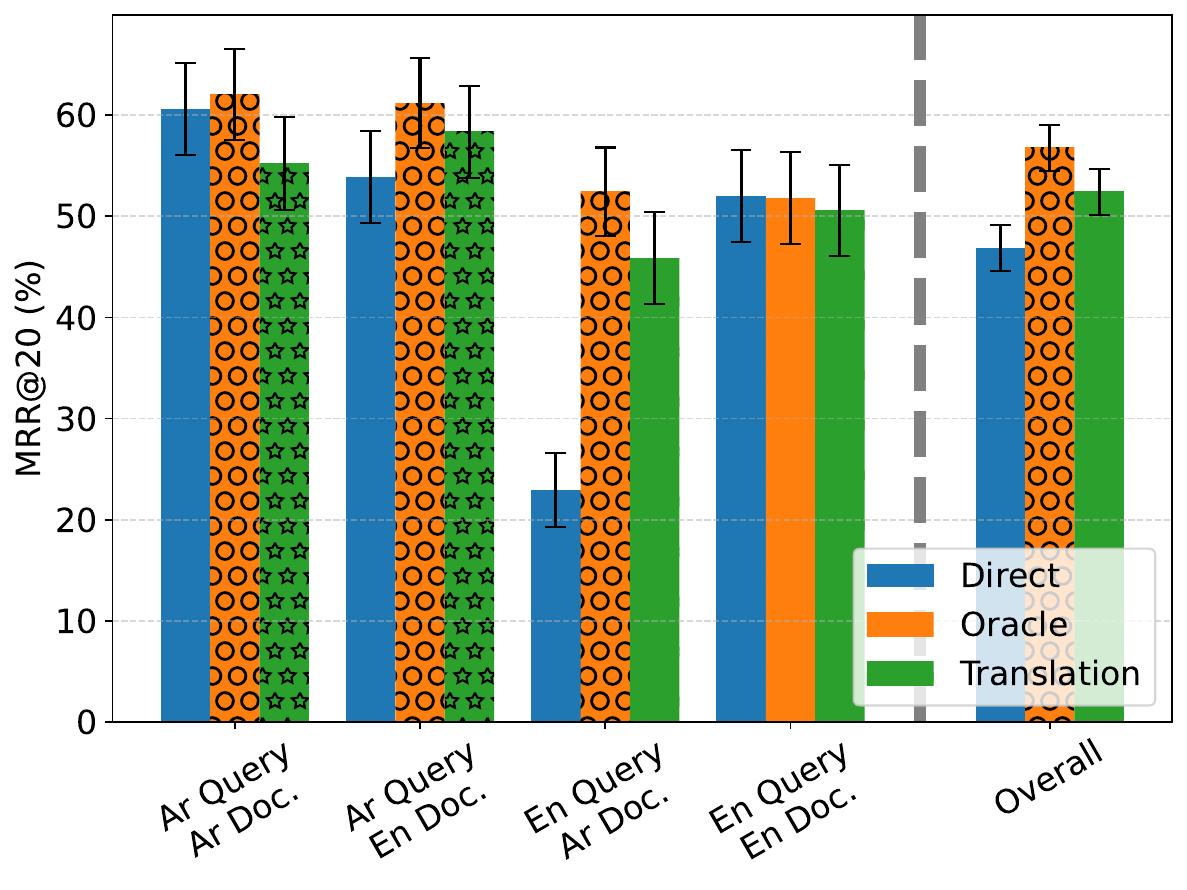}
      \label{fig:mrr:legal:bge}
    \end{subfigure} &
    \begin{subfigure}[t]{0.45\textwidth}
      \centering
      \caption{Legal benchmark – \Efive embedder}
      \includegraphics[width=\linewidth]{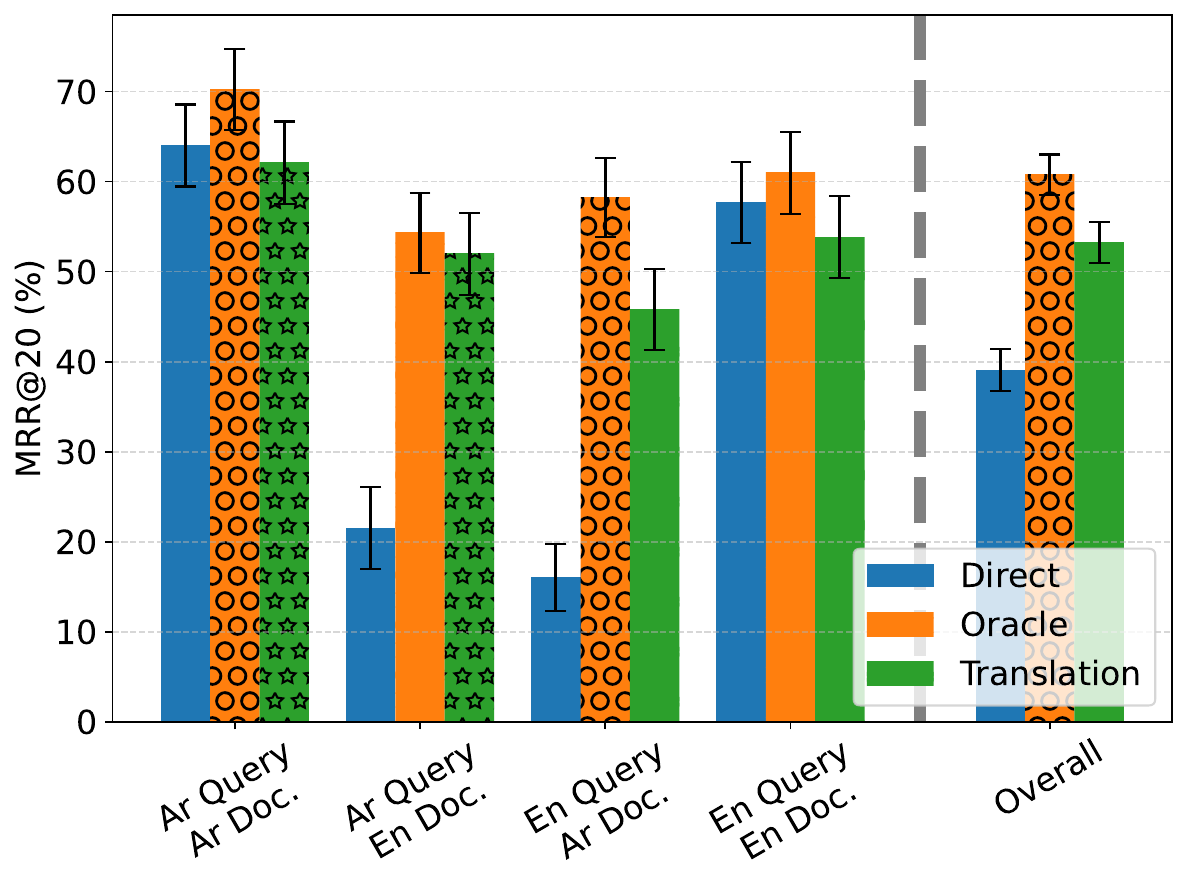}
      \label{fig:mrr:legal:e5}
    \end{subfigure} \\

    \begin{subfigure}[t]{0.45\textwidth}
      \centering
      \caption{Travel benchmark – \BGE embedder}
      \includegraphics[width=\linewidth]{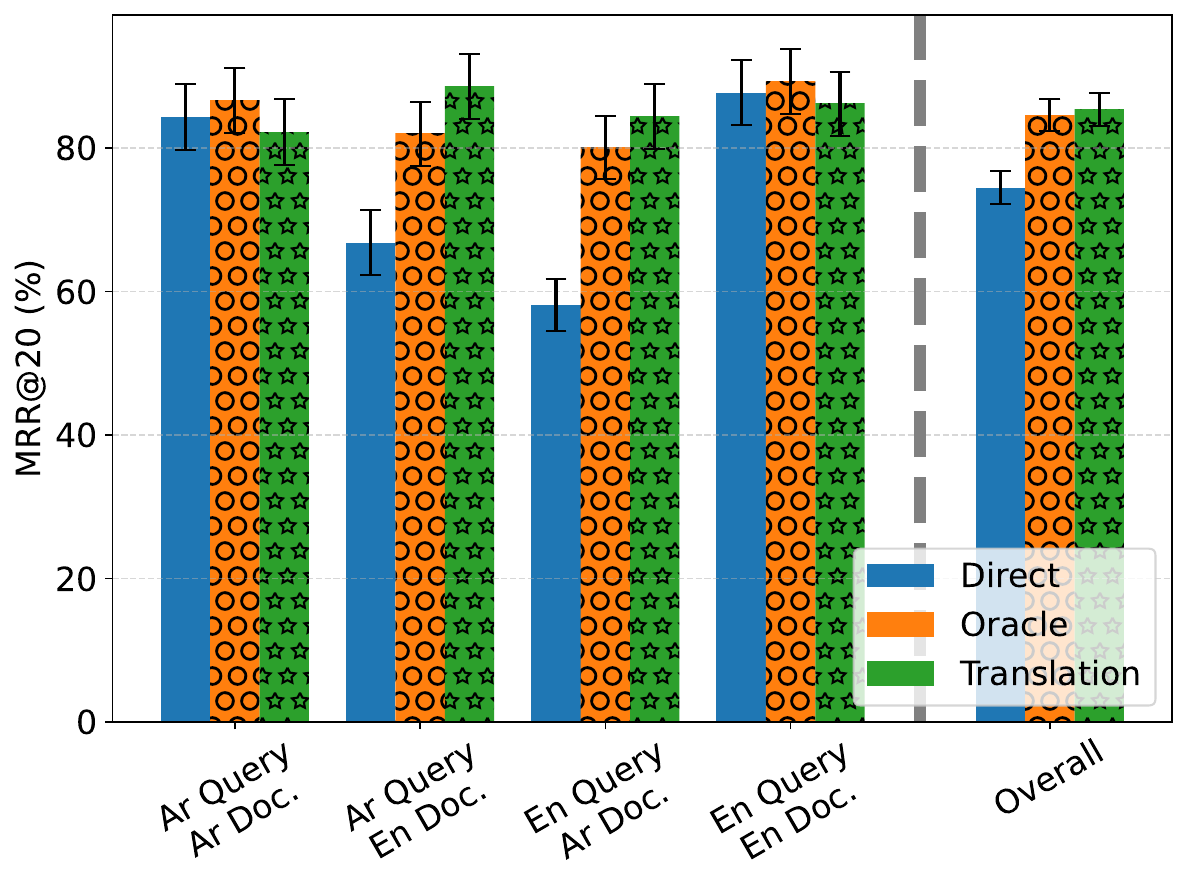}
      \label{fig:mrr:mofa:bge}
    \end{subfigure} &
    \begin{subfigure}[t]{0.45\textwidth}
      \centering
      \caption{Travel benchmark – \Efive embedder}
      \includegraphics[width=\linewidth]{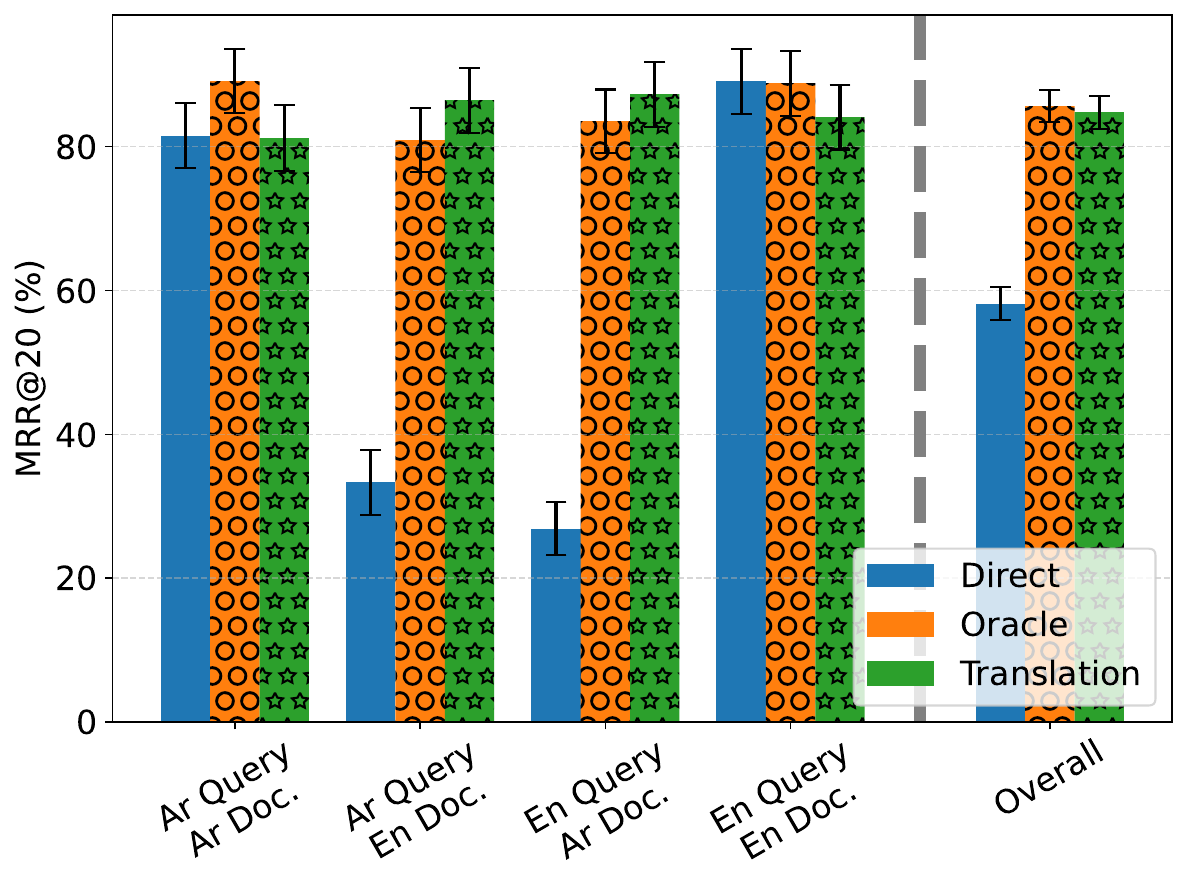}
      \label{fig:mrr:mofa:e5}
    \end{subfigure}
  \end{tabular}

  \caption{{\bf Retrieval MRR@20 scores across benchmarks and embedders.} Each figure corresponds to a specific combination of benchmark and embedding.
Bars represent retrieval MRR@20 scores in percentages, with 95\% confidence intervals shown as black error lines.
Different retrieval policies are distinguished by color and texture.
Results are grouped by benchmark segments defined by the user-document language combination, as well as the overall benchmark retrieval performance.}
  \label{fig:mrr:all}
\end{figure*}

Table~\ref{tab:app:per-language} provides the counterpart to Table~\ref{tab:per-language}, augmented with MRR@20 and NDCG@20 results. Figures~\ref{fig:ndcg:all} and \ref{fig:mrr:all} present a variation of Figure~\ref{fig:hit20:all} but according to NDCG@20 and MRR@20 respectively. The results for the balanced retriever are omitted because ranking-based metrics like NDCG@20 and MRR@20 require a single, consistent ordering of retrieved passages. Since the balanced retriever returns two separate rank lists (one in Arabic and one in English), these metrics cannot be meaningfully computed.
As can be seen, consistent trends occur for all metrics: Cross-language performance is worse compared to same-language performance. In fact, the gap in most scenarios is more pronounced for NDCG@20 and MRR@20 compared to Hit@20.

\begin{table*}[ht]
  \centering
  \begin{minipage}{0.48\textwidth}
\centering
\resizebox{\linewidth}{!}{%
\begin{tabular}{c|cc|ccc}
\toprule
\textbf{Benchmark} & \makecell{\textbf{User} \\ \textbf{Lang.}} & \makecell{\textbf{Doc} \\ \textbf{Lang.}} & \makecell{\textbf{Hit}\\ \textbf{@20}} & \makecell{\textbf{NDCG}\\ \textbf{@20}} & \makecell{ \textbf{MRR}\\ \textbf{@20}} \\
\hline
\multirow{6}{*}{Legal}  & arabic & arabic  & 92.4 & 65.9 & 60.6 \\
                          & arabic & english  & 89.8 & 60.6 & 53.9 \\
                          & english & arabic  & 55.8 & 29.9 & 22.9 \\
                          & english & english & 86.5 & 58.7 & 52.0 \\
  \cdashline{2-6}
                          & same-lang. &   & 89.4 & 62.3 & 56.3 \\
                          & cross-lang. &  & 72.8 & 45.3 & 38.4 \\
\hline
\multirow{6}{*}{Travel}  & arabic & arabic  & 92.8 & 85.5 & 84.4 \\
                       & arabic & english & 91.0 & 71.7 & 66.8 \\
                       & english & arabic & 80.0 & 62.5 & 58.1 \\
                       & english & english & 93.6 & 88.3 & 87.8 \\
  \cdashline{2-6}
                       & same-lang. &  & 93.2 & 86.9 & 86.1 \\
                       & cross-lang. &  & 85.5 & 67.1 & 62.5 \\
\hline
\bottomrule
\end{tabular}
}
\subcaption{{ \BGE embedder} }
\label{tab:app:per-language-bge}
  \end{minipage}%
  \hfill
  \begin{minipage}{0.48\textwidth}
    \centering
\resizebox{\linewidth}{!}{%
\begin{tabular}{c|cc|cccc}
\toprule
\textbf{Benchmark} & \makecell{\textbf{User} \\ \textbf{Lang.}} & \makecell{\textbf{Doc} \\ \textbf{Lang.}} & \makecell{\textbf{Hit}\\ \textbf{@20}} & \makecell{\textbf{NDCG}\\ \textbf{@20}} & \makecell{ \textbf{MRR}\\ \textbf{@20}} \\
\hline
\multirow{6}{*}{Legal}  & arabic & arabic & 87.5 & 66.4 & 64.0 \\
                          & arabic & english & 50.6 & 27.4 & 21.5 \\
                          & english & arabic & 40.7 & 21.4 & 16.1 \\
                          & english & english  & 87.7 & 62.8 & 57.7 \\
  \cdashline{2-7}
                          & same-lang. &  & 87.6 & 64.6 & 60.8 \\
                          & cross-lang. &  & 45.6 & 24.4 & 18.8 \\
\hline
\multirow{6}{*}{Travel}  & arabic & arabic & 90.0 & 82.8 & 81.5 \\
                       & arabic & english & 54.3 & 37.4 & 33.3 \\
                       & english & arabic & 63.9 & 35.1 & 26.9 \\
                       & english & english & 94.9 & 89.5 & 89.1 \\
  \cdashline{2-7}
                       & same-lang. & & 92.5 & 86.2 & 85.3 \\
                       & cross-lang. &  & 59.1 & 36.3 & 30.1 \\
\hline
\bottomrule
\end{tabular}
}
\subcaption{{ \Efive embedder} 
}
\label{tab:app:per-language-e5}
  \end{minipage}
\caption{{\bf Retriever Performance across language combinations.} 
Retriever performance is reported using three metrics: Hit@20, NDCG@20, and MRR@20.
Results are presented for each embedder, benchmark and for each of the four possible user–document language combinations. In addition, we report same-language and cross-language scores, defined as the mean scores over combinations where the user and document languages match or differ, respectively.
}
\label{tab:app:per-language}
\end{table*}

\subsection{Imbalanced corpora}\label{app:imbalanced}
In this section, we explore imbalanced corpora where one language dominates. We added experiments with different bilingual corpus ratios to examine whether the observed trends persist.

Given a target fraction $X$ of English documents, we construct a corpus in the same manner as the original one in the paper, but retain the English version of each document with probability $X$. The existing corpus corresponds to $X=50\%$, and we added two new corpora for $25\%$ and $75\%$. We evaluated three methods: (i) the direct approach, (ii) the balanced approach that retrieves 10 documents from each language, denoted Balanced-Equal, and (iii) a new method we call Balanced-Weighted, which retrieves English and Arabic documents in proportion to their ratio in the corpus (e.g., the top-$5$ documents in English and the top-$15$ in Arabic for $X=25\%$). The results are presented in Table~\ref{tab:uneven_corpus_travel}.

\begin{table*}[ht]
  \centering
\begin{subtable}{\textwidth}
    \centering
\begin{tabular}{lllll}
\toprule
User-Doc Langs. & Retriever & 25\% English & 50\% English & 75\% English \\
\midrule
Same-lang. & Direct & 95±1\% & 93±2\% & 92±2\% \\
Same-lang. & Balanced (Equal) & 95±1\% & 95±2\% & 93±1\% \\
Same-lang. & Balanced (Weighted) & 96±1\% & 95±2\% & 93±1\% \\
Cross-lang. & Direct & 79±2\% & 86±3\% & 86±2\% \\
Cross-lang. & Balanced (Equal) & 89±2\% & 93±2\% & 94±2\% \\
Cross-lang. & Balanced (Weighted) & 90±2\% & 93±2\% & 92±2\% \\
\bottomrule
\end{tabular}
\caption{\BGE Embedder}
\end{subtable}
  
\begin{subtable}{\textwidth}
    \centering
\begin{tabular}{lllll}
\toprule
User-Doc Langs. & Retriever & 25\% English & 50\% English & 75\% English \\
\midrule
Same-lang. & Direct & 96±1\% & 92±2\% & 92±2\% \\
Same-lang. & Balanced-Equal & 96±1\% & 96±2\% & 94±1\% \\
Same-lang. & Balanced-Weighted & 96±1\% & 96±2\% & 94±1\% \\
Cross-lang. & Direct & 65±3\% & 59±4\% & 56±3\% \\
Cross-lang. & Balanced-Equal & 91±2\% & 94±2\% & 93±2\% \\
Cross-lang. & Balanced-Weighted & 93±2\% & 94±2\% & 92±2\% \\
\bottomrule
\end{tabular}
\caption{\Efive Embedder}
\end{subtable}

\caption{Hit@20 scores across different corpus imbalances for the Travel benchmark}
\label{tab:uneven_corpus_travel}
\end{table*}

We draw 2 notable conclusions from Table~\ref{tab:uneven_corpus_travel}: (1) The Balanced-Equal baseline is stable in its performance across the 3 corpora, up to statistical noise. (2) The improvement of Balanced-Equal compared to the Direct baseline, as well as its competitiveness when compared to Balanced-Weighted remain across all settings, including the imbalanced corpora.

\end{document}